# AKILLI TELEFON SENSÖR VERİLERİ KULLANILARAK İNSAN AKTİVİTESİ TANIMA İÇİN XGBOOST VE MİNİROCKET ALGORİTMALARININ KARŞILAŞTIRMALI ANALİZİ

Dr. Celal Alagöz
ORCID: 0000-0001-9812-1473
celal.alagoz@gmail.com
Bağımsız, Kırıkkale, Türkiye

**Özet**

İnsan aktivitelerinin tanınması literatürde geniş kapsamlı bir şekilde incelenmekle birlikte, son zamanlarda doğru sınıflandırma için gelişmiş makine öğrenme ve derin öğrenme algoritmalarının uygulanmasına odaklanılmıştır. Bu çalışma, Extreme Gradient Boosting ve MiniRocket olmak üzere iki makine öğrenimi algoritmasının etkinliğini, akıllı telefon sensörlerinden elde edilen verilerle insan aktivitesi tanıma alanında araştırmaktadır. Deneyler, akıllı telefon giyen 30 gönüllünün çeşitli aktiviteleri gerçekleştirirken kaydedilen ivmeölçer ve jiroskop sinyallerini içeren, University of California Irvine açık erişim deposundan bir veri kümesi üzerinde gerçekleştirilir. Veri kümesi, eğitim ve test verileri için kullanılmadan önce gürültü filtreleme ve özellik çıkarma gibi ön işlemlere tabi tutulmuştur. Modelin sağlamlığını değerlendirmek için Monte Carlo çapraz doğrulama ve başarı değerlendirmesi için doğruluk, F1 skoru ve eğri altındaki alan gibi metrikler kullanıldı. Sonuçlar, hem Extreme Gradient Boosting'in hem de MiniRocket'in aktiviteleri sınıflandırmada doğruluk, F1 skoru ve eğri altındaki alan değerlerinin 0.99'a kadar ulaştığını göstermektedir. Extreme Gradient Boosting, MiniRocket'e kıyasla hafifçe üstün bir performans sergilemektedir. Elde edilen performans, literatürde insan aktivitesi tanıma görevleri için bildirilen diğer makine öğrenmesi ve derin öğrenme algoritmalarının performansını aşmaktadır. Ek olarak, çalışma iki algoritmanın hesaplama verimliliğini karşılaştırır ve Extreme Gradient Boosting'in eğitim süresi açısından avantajlı olduğunu ortaya koyar. Ayrıca, Minirocket'in performansı, ham verileri kullanarak ve sensörlerden yalnızca bir kanalı kullanarak doğruluk ve F1 değerlerini 0.94, eğri altındaki alan değerini ise 0.96 olarak elde etmesi, işlenmemiş sinyalleri doğrudan kullanma potansiyelini vurgulamakla birlikte, sensör füzyonu veya kanal füzyonu tekniklerinden yararlanılarak elde edilebilecek potansiyel avantajları işaret eder. Genel olarak, bu araştırma, Extreme Gradient Boosting ve MiniRocket'in insan aktivitesi tanıma görevlerindeki etkililiğini ve hesaplama özelliklerini aydınlatarak, akıllı telefon sensör verilerini kullanarak aktivite tanıma alanında gelecek çalışmalar için görüşler sunar.

**Anahtar Kelimeler: İnsan Aktivitesi Tanıma, Zaman Serisi Sınıflandırma, Makine Öğrenmesi.**

# COMPARATIVE ANALYSIS OF XGBOOST AND MINIROCKET ALGORITHMS FOR HUMAN ACTIVITY RECOGNITION USING SMARTPHONE SENSOR DATA

**Abstract**

Human activity recognition has been extensively studied, with recent emphasis on the implementation of advanced machine learning and deep learning algorithms for accurate classification. This study investigates the efficacy of two machine learning algorithms, Extreme Gradient Boosting and MiniRocket, in the realm of human activity recognition using data collected from smartphone sensors. The experiments are conducted on a dataset obtained from the University of California Irvine repository, comprising accelerometer and gyroscope signals captured from 30 volunteers performing various activities while wearing a smartphone. The dataset undergoes preprocessing, including noise filtering and feature extraction, before being utilized for training and testing the classifiers. Monte Carlo cross-validation is employed to evaluate the models' robustness, with metrics such as accuracy, F1 score, and area under the curve utilized for performance assessment. The findings reveal that both Extreme Gradient Boosting and MiniRocket attain accuracy, F1 score, and area under the curve values as high as 0.99 in activity classification. Extreme Gradient Boosting exhibits a slightly superior performance compared to MiniRocket. Notably, both algorithms surpass the performance of other machine learning and deep learning algorithms reported in the literature for human activity recognition tasks. Additionally, the study compares the computational efficiency of the two algorithms, revealing Extreme Gradient Boosting's advantage in terms of training time. Furthermore, the performance of MiniRocket, which achieves accuracy and F1 values of 0.94, and an area under the curve value of 0.96 using raw data and utilizing only one channel from the sensors, highlights the potential of directly leveraging unprocessed signals. It also suggests potential advantages that could be gained by utilizing sensor fusion or channel fusion techniques. Overall, this research sheds light on the effectiveness and computational characteristics of Extreme Gradient Boosting and MiniRocket in human activity recognition tasks, providing insights for future studies in activity recognition using smartphone sensor data.

**Keywords:** Human Activity Recognition, Time Series Classification, Machine Learning.

## INTRODUCTION

Human activity recognition (HAR) enables machines to interpret and analyze a variety of human activities using input data sources such as sensors and multimedia content (Dang et al., 2019). The rapid advancement of smartphones, wearable devices, and closed-circuit television systems has spurred researchers to enhance HAR systems for practical applications. HAR finds application in surveillance systems (Jalal et al., 2017), behavior analysis (Batchuluun et al., 2017), gesture recognition (Pigou et al., 2018), patient monitoring systems (Capela et al., 2015), ambient assisted living (Sankar et al., 2018), and various healthcare systems (Qi et al., 2018) that involve direct or indirect interaction between humans and smart devices.

For a considerable period, various traditional machine learning (ML) algorithms have been employed to address the HAR challenge. Meanwhile, deep learning (DL) has garnered considerable interest within the community due to its outstanding performance across several domains, including image (Obaid et al., 2020), audio (Zaman et al., 2023), and text (Minaee) data. A notable feature of DL is its ability to significantly reduce the effort required for feature selection compared to traditional ML algorithms, achieving this by automatically extracting abstract features through multiple hidden layers. Consequently, there has been a proliferation of DL-based HAR frameworks, which have been introduced recently (Dang et al., 2020). Given the popularity of HAR, numerous studies have been conducted to survey and compare deep learning and machine learning techniques (Dang et al., 2020; Ye et al., 2020).

While traditional ML algorithms have shown impressive performance with limited training data, they require numerous preprocessing steps and carefully engineered features, leading to inefficiencies and time consumption. On the other hand, deep learning techniques do not require manual feature crafting. However, they are resource intensive requiring both large data and extensive computational power to achieve a considerable improvement in efficacy, making them hard to apply in case of limited data and cases that a quicker analysis is required. Also, they are not readily effective in all topics and data types (Shwartz-Ziv and Armon, 2022).

In the verge of this trade-off, this study conveys an experimental analysis by running two ML algorithms on the HAR problem. One of them is Extreme Gradient Boosting (XGBoost), a machine learning algorithm of the gradient boosted decision tree type, introduced by Chen and Guestrin in 2016. XGBoost has demonstrated state-of-the-art performance on numerous tabular datasets, as highlighted by studies such as Ramraj et al. (2016) and Zhao et al. (2019). The other is MiniRocket, a time series classification algorithm developed by Dempster et al. 2020 that efficiently and effectively exploits convolution operation, has seamless implementation without requiring extensive preprocessing steps or feature extraction, minimal parameter tuning requirements, and superior adaptability to diverse signal sizes.

In summary, this study presents an experimental analysis comparing two ML algorithms for HAR. XGBoost and MiniRocket are evaluated on a dataset from the University of California Irvine (UCI) repository, comprising accelerometer and gyroscope signals from 30 volunteers. The dataset undergoes preprocessing before being used for training and testing the classifiers. Both algorithms achieve high accuracy, F1 score, and area under the curve (AUC) values, with XGBoost showing slightly superior performance. Notably, both algorithms outperform other ML and DL algorithms reported in the literature for HAR tasks. Furthermore, MiniRocket's performance highlights the potential of directly leveraging unprocessed signals and suggests potential advantages of utilizing sensor fusion or channel fusion techniques. Overall, this research provides insights into the effectiveness and computational characteristics of XGBoost and MiniRocket in HAR tasks, guiding future studies in smartphone sensor data-based activity recognition.

## MATERIALS AND METHODS

**Dataset Description:**

The experiment utilizes a public dataset available on the UCI website, involving a cohort of 30 volunteers aged between 19 and 48 years. Each participant engaged in six distinct activities—walking, walking upstairs, walking downstairs, sitting, standing, and laying—while wearing a Samsung Galaxy S II smartphone on their waist. Through the smartphone's embedded accelerometer and gyroscope, testers captured 3-axial linear acceleration and 3-axial angular velocity at a consistent rate of 50Hz. To ensure accuracy, the experiments were video-recorded for manual data labeling.

Prior to analysis, the sensor signals (accelerometer and gyroscope) underwent preprocessing, which included noise filtering. Subsequently, the signals were sampled using fixed-width sliding windows of 2.56 seconds with a 50% overlap (128 readings per window). Through a Butterworth low-pass filter, the sensor acceleration signal was decomposed into body acceleration and gravity components, with the latter assumed to contain only low-frequency elements, thus utilizing a filter with a 0.3 Hz cutoff frequency.

From each window, a feature vector was derived by computing variables from both the time and frequency domains. Each dataset record includes triaxial acceleration data from the accelerometer (representing total acceleration and estimated body acceleration), triaxial angular velocity data from the gyroscope, a 561-feature vector comprising time and frequency domain variables, the corresponding activity label, and an identifier for the subject who conducted the experiment.

**Precomputed Features:**

The features incorporated into this database originate from the raw signals of the accelerometer and gyroscope, which consist of 3-axial measurements. Subsequently, they underwent filtering procedures involving a median filter and a 3rd order low pass Butterworth filter with a corner frequency of 20 Hz to eliminate noise. Additionally, the acceleration signal was segmented into body and gravity components using another low pass Butterworth filter with a corner frequency of 0.3 Hz.

Following this, the body linear acceleration and angular velocity were computed over time to generate jerk signals. Furthermore, the magnitude of these three-dimensional signals was determined using the Euclidean norm. Subsequent to these computations, a Fast Fourier Transform was applied to certain signals, resulting in frequency domain representations.

These signals were instrumental in estimating various variables constituting the feature vector for each pattern. The variables derived from these signals include: Mean value, Standard deviation, Median absolute deviation, Maximum value in array, Minimum value in array, Signal magnitude area, Energy measure, calculated as the sum of squares divided by the number of values, Interquartile range, Signal entropy, Autoregression coefficients with Burg order equal to 4, Correlation coefficient between two signals, Index of the frequency component with the largest magnitude, Weighted average of the frequency components to obtain a mean frequency, Skewness of the frequency domain signal, Kurtosis of the frequency domain signal, Energy of a frequency interval within the 64 bins of the FFT of each window, Angle between two vectors.

**XGBoost:**

XGBoost, developed by Tianqi Chen and collaborators, stands as an open-source gradient boosting framework widely recognized for its exceptional performance and adaptability, particularly in handling structured/tabular data. Below are its salient features and attributes:

**Gradient Boosting**: XGBoost is a member of the ensemble learning techniques, specifically the gradient boosting algorithms. It sequentially builds multiple weak learners, with each subsequent learner focusing on rectifying the errors of its predecessors.

**Tree Ensemble Method**: XGBoost predominantly employs decision trees as base learners, forming an ensemble of decision trees where each tree aims to rectify the errors of the preceding trees in the ensemble.

**Regularization**: To prevent overfitting, XGBoost incorporates regularization techniques by including both L1 (Lasso) and L2 (Ridge) regularization terms in its objective function. This effectively controls the model's complexity and enhances generalization performance.

**Scalability**: Engineered for efficiency and scalability, XGBoost efficiently handles large datasets with millions of instances and features. It is optimized for speed and memory usage, enabling swift model training and prediction.

**Handling Missing Values**: XGBoost natively supports handling missing values in input data. It autonomously learns how to manage missing values during training without requiring preprocessing.

**Flexibility**: XGBoost offers support for various objective functions and evaluation metrics, making it adaptable to diverse supervised learning tasks such as classification, regression, and ranking.

**Feature Importance**: XGBoost provides feature importance scores, revealing the contribution of each feature to the model's predictions. This aids users in understanding the relative significance of different features in their datasets.

**Community Support**: With a large and active community of users and contributors, XGBoost undergoes continuous enhancements, bug fixes, and extensions. Widely adopted across various industries, it is frequently the algorithm of choice in data science competitions.

In summary, XGBoost emerges as a potent and versatile machine learning algorithm, excelling in predictive modeling tasks involving structured/tabular data. Its blend of performance, scalability, and flexibility positions it as a preferred tool for data scientists and machine learning practitioners worldwide.

**Minirocket:**

MiniRocket is a state-of-the-art lightweight time series classification algorithm introduced by Dempster et al. in 2020. It stands out for its remarkable efficiency and high accuracy in analyzing time series data, making it suitable for applications with large datasets or constrained computational resources.

Key features of MiniRocket include:

**Efficiency**: MiniRocket achieves exceptional computational efficiency by utilizing random convolutional kernels and the hashing trick. This enables it to efficiently process time series data without the need for complex model architectures or extensive computational resources.

**Scalability**: MiniRocket is designed to scale well with large datasets, making it suitable for real-world applications where handling massive amounts of time series data is essential.

**Accuracy**: Despite its lightweight design, MiniRocket delivers competitive classification accuracy compared to more complex and resource-intensive algorithms. It achieves this by leveraging randomized feature extraction techniques tailored specifically for time series classification tasks.

**Versatility**: MiniRocket's versatility allows it to handle various types of time series data, including univariate and multivariate data, making it applicable across a wide range of domains and applications.

Overall, MiniRocket offers a compelling solution for time series classification tasks, combining efficiency, scalability, and accuracy in a lightweight framework. Its introduction has significantly contributed to advancing the field of time series analysis and classification.

**Performance Evaluation:**

Monte Carlo cross-validation, a technique involving the random partitioning of the dataset into training and testing sets across multiple iterations, is utilized to ensure robust cross-validation. In this experimental setup, various combinations of training and testing sizes are explored to comprehensively assess classifier performance. To guarantee the reproducibility of training and testing sets, the shuffling seed for each iteration is determined by the iteration number.

**Accuracy**: This metric gauges the overall correctness of the classifier by determining the ratio of correctly predicted instances to the total instances, offering an intuitive assessment of its performance.

**F1 Score**: The F1 score is a widely used metric for evaluating classifier performance in binary classification tasks. It calculates the harmonic mean of precision and recall, providing a balanced measure of the classifier's ability to correctly identify positive instances while minimizing false positives and false negatives. In multi-class scenarios, the study adopts the macro-averaging approach, where the F1 score is independently computed for each class and then averaged across all classes, with equal weight assigned to each class.

**AUC**: The AUC is a metric used to assess the performance of binary classification models. It represents the area under the Receiver Operating Characteristic (ROC) curve, which plots the true positive rate (sensitivity) against the false positive rate (1 - specificity) at various threshold settings. A higher AUC value indicates superior discrimination ability of the model. In multi-class classification tasks, the concept of AUC can be extended using various methods such as One-vs-One (OvO), One-vs-All (OvA), macro-averaging, and weighted-averaging AUC. This study adopts the OvO approach where pairwise comparisons are made between all pairs of classes. For each pair of classes, a binary classifier is trained to distinguish between the instances of those two classes. The AUC is then calculated for each pair of classes, and the average AUC across all pairs is computed as the final multi-class AUC.

Table 1. Classification results usng different classifier and inputs models.

| Model | Accuracy | F1 | AUC | Training Time (sec) |
|---|---|---|---|---|
| *Precomputed Features* | | | | |
| XGBoost | 0.9896 ± 0.0022 | 0.9896 ± 0.0022 | 0.9999 ± 0.0000 | 26.9 |
| Minirocket | 0.9881 ± 0.0031 | 0.9886 ± 0.0029 | 0.9932 ± 0.0018 | 80.1 |
| *Total Acceleration (Minirocket)* | | | | |
| x-axis | 0.9040±0.0050 | 0.9088±0.0051 | 0.9454±0.0030 | 73.6 |
| y-axis | 0.9350±0.0054 | 0.9388±0.0051 | 0.9633±0.0031 | 73.8 |
| z-axis | 0.8721±0.0039 | 0.8801±0.0036 | 0.9282±0.0022 | 73.8 |
| *Body Gyro (Minirocket)* | | | | |
| x-axis | 0.8184±0.0073 | 0.8249±0.0073 | 0.8951±0.0043 | 74.1 |
| y-axis | 0.7940±0.0052 | 0.8065±0.0049 | 0.8844±0.0029 | 74.0 |
| z-axis | 0.8046±0.0056 | 0.8157±0.0058 | 0.8902±0.0033 | 74.5 |
| *Body Acceleration (Minirocket)* | | | | |
| x-axis | 0.8111 ± 0.0089 | 0.8232 ± 0.0086 | 0.8945 ± 0.0051 | 73.1 |
| y-axis | 0.7791 ± 0.0057 | 0.7922 ± 0.0055 | 0.8765 ± 0.0033 | 73.1 |
| z-axis | 0.7399 ± 0.0072 | 0.7565 ± 0.0069 | 0.8550 ± 0.0040 | 73.0 |

The experiments are carried out on a system featuring an Intel(R) Core(TM) i7-5820K CPU @ 3.30GHz x 11 and 16 GB of RAM. The computations are executed using the Python programming language.

**RESULTS AND DISCUSSIONS**

The HAR problem has garnered significant attention in recent years, leading to the development of various ML and deep learning DL algorithms aimed at addressing classification challenges. This study introduces two ML models, previously unexplored in this specific problem domain, namely Minirocket and XGBoost. Minirocket is a cutting-edge, lightweight time series classification algorithm, while XGBoost is a widely acclaimed ML algorithm renowned for its success in handling tabular data and adeptness at managing class imbalances.

Two classification approaches were employed: utilizing precomputed features provided with the dataset and utilizing raw time series data from a single channel. The latter approach exclusively utilized Minirocket. Table 1 presents the classification results, demonstrating near-perfect performance in terms of accuracy, F1 score, and AUC for both models. XGBoost marginally outperforms Minirocket with 0.99 accuracy, F1 score, and AUC. However, considering training time, XGBoost significantly outshines Minirocket, requiring only 26.9 seconds compared to 80.1 seconds, thus establishing itself as the preferred model for this task.

Further insight is provided through confusion matrices depicted in Figure 1. Notably, Minirocket primarily confuses between 'sitting' and 'standing,' exhibiting balanced false positives and false negatives. Conversely, XGBoost displays similar confusion between 'sitting' and 'standing,' with slight confusions observed among 'walking,' 'walking upstairs,' and 'walking downstairs' activities.

The exceptional performance of XGBoost, given its proficiency with tabular data, is unsurprising. However, the notable performance of Minirocket with tabular data is unexpected, showcasing its versatility beyond its primary domain of time series data analysis, thereby opening avenues for further research.

Another classification scheme employed triaxial inertial signals: acceleration from the accelerometer (total acceleration), angular velocity from the gyroscope, and estimated body acceleration, resulting in a multivariate time series data with 9 channels. In this approach, Minirocket operated in a univariate manner, analyzing each channel individually without requiring preprocessing steps. While generally exhibiting lower performance compared to using precomputed features, notable results were achieved, particularly with the y-axis component of total acceleration, yielding 0.94 accuracy, F1 score, and 0.96 AUC.

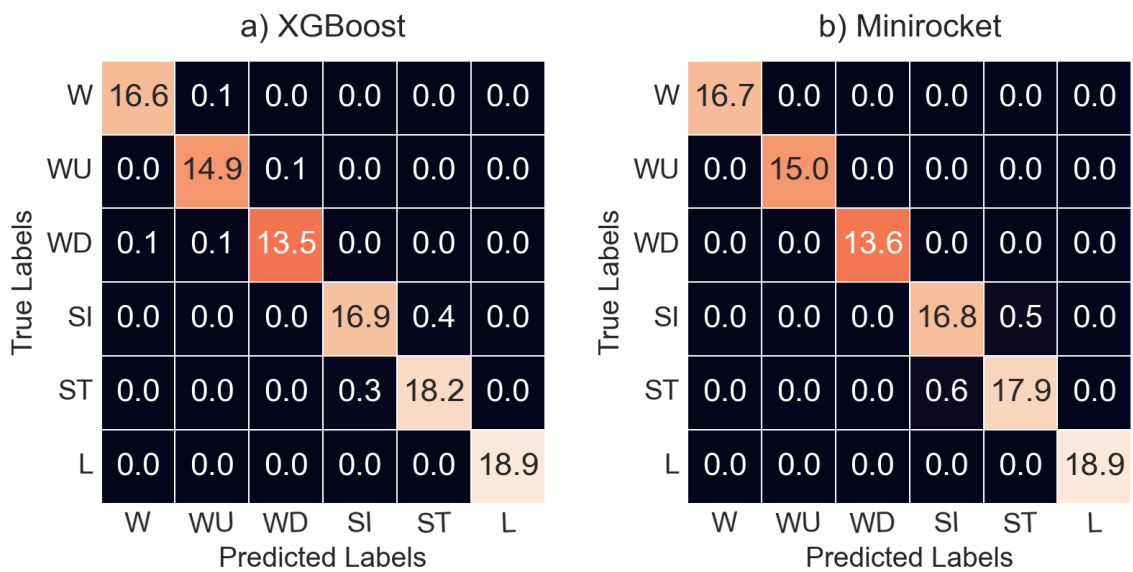

Figure 1. Confusion matrices resulting from 10 iterations of Monte Carlo validation for XGBoost and Minrocket algorithms employing precomputed features. Values within each cell denote the percentage of total test instances across all runs.

Table 2. Comparison with reported literature findings

| **Model** | **Accuracy** | **F1** | **AUC** |
|---|---|---|---|
| XGBoost | **0.990** | **0.990** | **0.999** |
| Minirocket | **0.988** | **0.989** | **0.993** |
| SGD | 0.446 | 0.427 | 0.664 |
| Naive Bayes | 0.736 | 0.747 | 0.734 |
| DT | 0.748 | 0.746 | 0.850 |
| kNN | 0.707 | 0.706 | 0.895 |
| RF | 0.818 | 0.818 | 0.966 |
| NN | 0.856 | 0.857 | 0.974 |
| SVM | 0.878 | **0.872** | **0.988** |
| LSTM | 0.900 | - | - |
| CNN | **0.975** | - | - |

Furthermore, in Table 2, the classification performances of the models in this study were compared with reported performances of several other algorithms from a recent study (Ye et al., 2020), including Stochastic Gradient Descent (SGD), Naive Bayes, Decision Tree (DT), k-Nearest Neighbor (kNN), Random Forest (RF), Neural Network (NN), Support Vector Machine (SVM), Long Short-Term Memory (LSTM), and Convolutional Neural Network (CNN). XGBoost consistently emerged as the top algorithm in terms of accuracy, F1 Score, and AUC, followed by Minirocket. CNN ranked third in accuracy, while SVM ranked third in F1 score and AUC.

Moving forward, future research should explore sensor fusion techniques and multivariate time series classification options to enhance classification performance in HAR tasks.

**CONCLUSIONS**

This study explored the effectiveness of two ML models, XGBoost and Minirocket, in addressing the HAR problem. Through comprehensive experimentation, several key findings and implications emerged:

**Model Performance**: Both XGBoost and Minirocket demonstrated remarkable performance in classifying human activities, achieving near-perfect accuracy, F1 score, and AUC. XGBoost exhibited marginally superior performance compared to Minirocket, particularly in terms of training efficiency.

**Versatility of Minirocket**: An unexpected observation was the robust performance of Minirocket with tabular data, despite its primary design for time series analysis. This versatility highlights Minirocket's potential applicability beyond its original scope, suggesting avenues for further research and exploration.

**Data Utilization Strategies**: The study compared classification performance using precomputed features and raw time series data, highlighting the advantages and trade-offs of each approach. While precomputed features yielded excellent results, Minirocket's performance with raw data underscores the potential of leveraging unprocessed signals directly.

**Comparison with Existing Algorithms**: The comparison with various state-of-the-art algorithms from a recent study reaffirmed the superior performance of XGBoost and Minirocket in the HAR domain. This further solidifies their standing as leading approaches for activity recognition tasks.

**Future Directions**: To enhance HAR classification accuracy, future research directions should focus on exploring sensor fusion techniques and multivariate time series classification methods. Additionally, investigating the generalizability of Minirocket across diverse datasets and domains could unveil its broader applicability.

In conclusion, this study contributes valuable insights into the application of XGBoost and Minirocket in HAR tasks, emphasizing their efficacy, versatility, and potential for further advancements in activity recognition research.